# A Novel Crossover Operator for Genetic Algorithms: Ring Crossover

Yılmaz KAYA[1]  Murat UYAR[2]  Ramazan TEKİN[3]

[1]Siirt University, Department of Computer Engineering, Siirt, Turkey, yilmazkaya1977@gmail.com
[2]Siirt University, Department of Electric and Electronics Engineering, Siirt, Turkey, muratuyar1@gmail.com
[3]Batman University, Department of Computer Engineering, Batman, Turkey, ramazan.tekin@batman.edu.tr

*Abstract*—The genetic algorithm (GA) is an optimization and search technique based on the principles of genetics and natural selection. A GA allows a population composed of many individuals to evolve under specified selection rules to a state that maximizes the "fitness" function. In that process, crossover operator plays an important role. To comprehend the GAs as a whole, it is necessary to understand the role of a crossover operator. Today, there are a number of different crossover operators that can be used in GAs. However, how to decide what operator to use for solving a problem? A number of test functions with various levels of difficulty has been selected as a test polygon for determine the performance of crossover operators.

In this paper, a novel crossover operator called 'ring crossover' is proposed. In order to evaluate the efficiency and feasibility of the proposed operator, a comparison between the results of this study and results of different crossover operators used in GAs is made through a number of test functions with various levels of difficulty. Results of this study clearly show significant differences between the proposed operator and the other crossover operators.

## I. Introduction

Genetic algorithms (GAs) represent general-purpose search and optimization technique based on evolutionary ideas of natural selection and genetics. They simulate natural processes based on principles of Lamarck and Darwin. In 1975, Holland developed this idea in his book "Adaptation in natural and artificial systems". He described how to apply the principles of natural evolution to optimization problems and built the first GAs. Holland's theory has been further developed and now GAs stand up as a powerful tool for solving search and optimization problems. GAs are based on the principle of genetics and evolution [1]. Today, there exists many variations on GAs and term "genetic algorithm" is used to describe concepts sometimes very far from Holland's original idea [2]. The two most commonly employed genetic search operators are crossover and mutation. Crossover produces offspring by recombining the information from two parents. Mutation prevents convergence of the population by flipping a small number of randomly selected bits to continuously introduce variation. The driving force behind GAs is the unique cooperation between selection, crossover and mutation operator. A genetic operator is a process used in GAs to maintain genetic diversity. The most widely used genetic operators are recombination, crossover and mutation.

The main goal of this paper is to introduce a new crossover operator called ring crossover (RC) and present the performance of this crossover operator. The rest of this paper is organized as follow. In section 2, definitions and concepts of the different crossover operators are introduced. In section 3, the proposed method in this study is given. In section 4, a number of the functions widely used in performance evaluation of GA operators are defined. In section 5, the optimization results and performance comparison of proposed method are shown. Finally, conclusions are discussed in section 6.

## II. Crossover operators

The crossover operator is a genetic operator that combines (mates) two chromosomes (parents) to produce a new chromosome (offspring). The idea behind crossover is that the new chromosome may be better than both of the parents if it takes the best characteristics from each of the parents. Crossover occurs during evolution according to a user-definable crossover probability. For purpose of this work, only crossover operators that operate on two parents and have no self-adaptation properties will be considered.

### A. Single Point Crossover

When performing crossover, both parental chromosomes are split at a randomly determined crossover point. Subsequently, a new child genotype is created by appending the first part of the first parent with the second part of the second parent [3, 4]. A single crossover point on both parents' organism strings is selected. All data beyond that point in either organism string is swapped between the two parent organisms. Figure 1 shows the single point crossover (SPC) process.

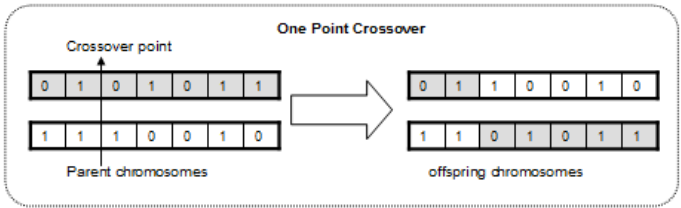

Figure 1. Single point crossover

B. *Two Point Crossover*

Apart from SPC, many different crossover algorithms have been devised, often involving more than one cut point. It should be noted that adding further crossover points reduces the performance of the GA. The problem with adding additional crossover points is that building blocks are more likely to be disrupted. However, an advantage of having more crossover points is that the problem space may be searched more thoroughly. In two-point crossover (TPC), two crossover points are chosen and the contents between these points are exchanged between two mated parents [5, 6]

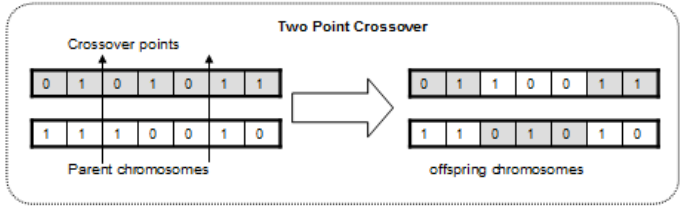

Figure 2. Two point crossover

In figure 2, the arrows indicate the crossover points. Thus, the contents between these points are exchanged between the parents to produce new children for mating in the next generation.

C. *Intermediate Crossover*

Intermediate creates offsprings by a weighted average of the parents. Intermediate crossover (IC) is controlled by a single parameter *Ratio*:

*offspring = parent1+rand\*Ratio\*(parent2 - parent1)*

If *Ratio* is in the range [0,1] then the offsprings produced are within the hypercube defined by the parents locations at opposite vertices. *Ratio* can be a scalar or a vector of length number of variables. If *Ratio* is a scalar, then all of the offsprings will lie on the line between the parents. If *Ratio* is a vector then children can be any point within the hypercube [7].

D. *Heuristic Crossover*

In heuristic crossover (HC), heuristic returns an offspring that lies on the line containing the two parents, a small distance away from the parent with the better fitness value in the direction away from the parent with the worse fitness value. The default value of *Ratio* is 1.2. If parent1 and parent2 are the parents, and parent1 has the better fitness value, the function returns the child [7],

*offspring = parent2 + Ratio \* (parent1 - parent2)*

E. *Arithmetic Crossover*

In arithmetic crossover (AC), arithmetic creates children that are the weighted arithmetic mean of two parents. Children are feasible with respect to linear constraints and bounds. Alpha is random value between [0,1]. If parent1 and parent2 are the parents, and parent1 has the better fitness value, the function returns the child [7],

*offspring =alpha\*parent1 + (1-alpha)\*parent2*

III. PROPOSED CROSSOVER OPERATOR: RING CROSSOVER

The operator called ring crossover is consisted of four steps. The steps of the proposed operator in this paper are shown in figure 3. All of the steps in the algorithm are discussed one by one.

*Step-1:* In this step, two parents such as parent1 and parent2 are considered for the crossover process, as shown in fig. 3(a).

*Step-2:* The chromosomes of parents are firstly combined with a form of ring, as shown in fig. 3(b). Later, a random cutting point is decided in any point of ring.

*Step-3:* The children are created with a random number generated in any point of ring according to the length of the combined two parental chromosomes. With reference to the cutting point in step 2, while one of the children is created in the clockwise direction, the other one is created in direction of the anti-clockwise, as shown in fig. 3(c).

*Step-4:* In this step, swapping and reversing process is performed in the RC operator, as shown in fig. 3(d). In swapping process, a number of genes are swapped in crossed parents. In reversion process, the remaining genes are reversed in crossed parents. As the length of ring is equal to the total length of both of parents and the children are created according to a random point of ring, more variety can be provided in possible number of children by RC operator according to SPC and TPC operators.

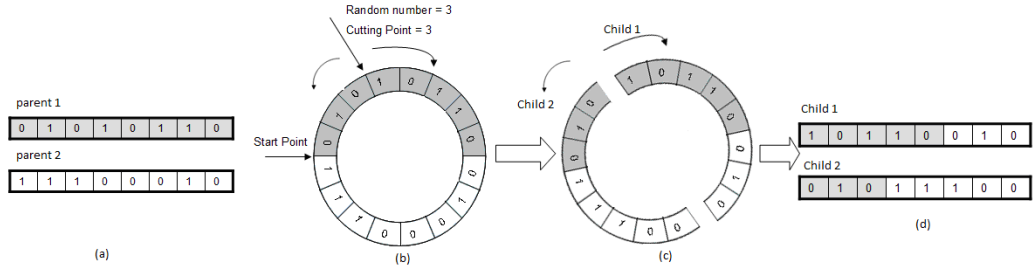

Figure 3. Ring crossover

## IV. TEST FUNCTIONS

The proposed method must be tested by a number of the functions widely used in performance evaluation of GA operators such as crossover. Test functions used in this paper have two important features: modality and separability. Unimodal function is a function with only one global optimum. Function is multimodal if it has two or more local optima. Multimodal functions are more difficult to optimize compared to unimodal functions [8].

### A. Sphere Function

Sphere function is a test function proposed by De Jong. It has been widely used in evaluation of genetic algorithms and development of the theory of evolutionary strategies. Sphere function is a simple, continuous and strongly convex function. Sphere function is unimodal and additively separable. Boundaries are set at [-5.12; 5.12]. Sphere function's global minimum is in point x=0 with value f(x)=0 [8,9]. The simplest test function is De Jong's function 1. It is continues, convex and unimodal. This function is defined as shown below.

$$f(x) = \sum_{i=1}^{D} x_i^2 \quad (1)$$

This function is defined as F1 in paper.

### B. Axis Parallel Hyper-Ellipsoid Function

This function is similar to Sphere function. It is also known as weighted sphere model. It is also unimodal and additively separable. Boundaries are set at [-5.12; 5.12]. Function's global minimum is in point x=0 with value f(x)=0 [8,9]. The axis parallel hyper-ellipsoid is similar to De Jong's function 1. It is also known as the weighted sphere model. Again, it is continues, convex and unimodal. It is defined as shown below.

$$f(x) = \sum_{i=1}^{D} i x_i^2 \quad (2)$$

This function is defined as F2 in paper.

### C. Rotated Hyper-Ellipsoid Function

This function represents an extension of the axis parallel hyper-ellipsoid function. With respect to the coordinate axes, this function produces rotated hyperellipsoids. It is continues, convex and unimodal. Boundaries are set at [-65.536; 65.536]. Function's global minimum is in point x=0 with value f(x)=0 [9]. An extension of the axis parallel hyper-ellipsoid is Schwefel's function 1.2. With respect to the coordinate axes, this function produces rotated hyper-ellipsoids. It is continues, convex and unimodal. This function is defined as shown below.

$$f(x) = \sum_{i=1}^{D} \left( \sum_{j=1}^{i} x_j \right)^2 \quad (3)$$

This function is defined as F3 in paper.

### D. Normalized Schwefel Function

The surface of Schwefel function is composed of a great number of peaks and valleys. The function has a second best minimum far from the global minimum where many search algorithms are trapped. Moreover, the global minimum is near the bounds of the domain. Schwefel's function is deceptive in that the global minimum is geometrically distant, over the parameter space, from the next best local minimum. Schwefel function is multimodal and additively separable. Boundaries are set at [-500; 500]. Function's global minimum is in point x=420.968 with value f(x)=-418.9829 [8] . Schwefel's function [Sch81] is deceptive in that the global minimum is geometrically distant, over the parameter space, from the next best local minima. Therefore, the search algorithms are potentially prone to convergence in the wrong direction. It is defined as shown below.

$$f(x) = \sum_{i=1}^{D} -x_i \sin(\sqrt{|x_i|}) \quad (4)$$

This function is defined as F4 in paper.

### E. Generalized Rastrigin Function

Rastrigin function was constructed from Sphere adding a cosine modular term. Its contour is made up of a large number of local minima whose value increases with the distance to the global minimum. Thus, the test function is highly multimodal. However, the location of the local minima's are regularly distributed. Rastrigin function is additively separable. Boundaries are set at [-5.12; 5.12]. Function's global minimum is found in point x=0 with value f(x)=0 [8,10].

$$f(x) = 10n - \sum_{i=1}^{D}(x_i^2 - 10\cos(2\pi x_i))  \quad (5)$$

This function is defined as F5 in paper.

*F. Rosenbrock's Valley Function*

Rosenbrock's valley function (banana function) is a classic optimization problem. The global optimum is inside a long, narrow, parabolic shaped flat valley. To find the valley is trivial, however convergence to the global optimum is difficult and hence this problem has been repeatedly used in assess the performance of optimization algorithms. Banana function is additively separable. Boundaries are set at [-2.048; 2.048]. Function's global minimum is found in point x=1 with value f(x)=0 [10]. This function is defined as shown below.

$$f(x) = \sum_{i=1}^{D-1} 100 - (x_{i+1} - x_i^2)^2 + (1 - x_i)^2 \quad (6)$$

This function is defined as F6 in paper.

## V. EXPERIMENTAL RESULTS

In all experiments, stochastic uniform selection was used. Parameters of GA for experiments were as following: Gaussian mutation with $p_m$ mutation coefficient of 0.01 and crossover rate $p_c$ of 0.8 was used, number of independent runs for each experiment was 30, initial population $N$ of size 20 was randomly created and used in experiments. Dimensionality of the search space $D$ for all test function was set to 30. Number of overall evaluations was set to 10000. For all test functions, finding global minimum is the objective. All of the experiment is realized for six different types of test functions. A comparison between the proposed crossover method (RC) and other crossover methods are made and the results are comparatively presented in table 1.

Table 1. Performance comparison for the different types of test functions

| Function | Results | SPC | TPC | IC | HC | AC | RC |
|---|---|---|---|---|---|---|---|
| F1 | Best | 5.732 | 3.416 | 6.207 | 0.011 | 5.589 | 0.0027 |
|  | Worst | 7.246 | 6.511 | 6.246 | 8.099 | 6.389 | 6.163 |
|  | Average | 5.737 | 3.417 | 6.208 | 2.81 | 5.589 | 0.3299 |
| F2 | Best | 70.52 | 68.63 | 64.04 | 0.024 | 73.71 | 0.1023 |
|  | Worst | 105.7 | 94.04 | 80.4 | 87.41 | 89.18 | 106.8 |
|  | Average | 70.52 | 68.64 | 64.04 | 5.706 | 73.72 | 11.73 |
| F3 | Best | 20.79 | 15.06 | 22.86 | 2.36 | 24.37 | 4.577 |
|  | Worst | 261.7 | 204.8 | 59.24 | 381 | 47.94 | 108.2 |
|  | Average | 37.02 | 16.22 | 22.87 | 17.58 | 24.37 | 18.97 |
| F4 | Best | -115.7 | -115.8 | -114.1 | -117.7 | -113.2 | -117.8 |
|  | Worst | -29.46 | -26.85 | -27.91 | -26.1 | -27.72 | -27.75 |
|  | Average | -115.6 | -115.4 | -114 | -117.1 | -113.1 | -117.7 |
| F5 | Best | 94.69 | 50.84 | 122.6 | 12.68 | 154 | 2.669 |
|  | Worst | 241.3 | 257.7 | 256.6 | 173.1 | 251.4 | 232.5 |
|  | Average | 111.3 | 52.15 | 187.3 | 31.98 | 154.1 | 3.691 |
| F6 | Best | 73.07 | 70.07 | 34.71 | 29.35 | 27.08 | 28.59 |
|  | Worst | 269.3 | 390.3 | 349.2 | 369.1 | 260.3 | 316.1 |
|  | Average | 73.08 | 78.39 | 34.74 | 117.5 | 27.12 | 32.69 |

## VI. CONCLUSION

In this paper, a new crossover operator called RC is proposed and experiments are conducted. The proposed operator is tested by a number of test functions with various levels of difficulty. A comparison between the results of this method and the results of other crossover operators are made. RC operator gives better results according to other crossover operators. Although the most of crossover operators showed similar results, RC operator had slightly better results than the other crossover for F1, F2, F4, F5 functions. For F3 function, HC operator has slightly better result than RC operator. However, RC operator produces better result than SPC, TPC, IC and AC. For F6 function, the results of this study are very close to those of AC, but in generally RC operator performed the best results than other crossover operators.

The most important advantage of the proposed method is that more variety is presented in possible number of children according to SPC and TPC operators. The experiments and the results presented in the paper clearly reveal the potential capability of the proposed method in optimization processing based on GA. Moreover, it has the great potential to improve the performance of GA applications in different area of engineering.